\pdfoutput=1

\documentclass[11pt]{article}

\usepackage[final]{ACL2023}

\usepackage{times}
\usepackage{latexsym}
\usepackage{tabularray}
\usepackage{tabularx}
\usepackage{makecell}
\usepackage{hyperref}
\usepackage{multirow}
\usepackage{multicol}
\usepackage{rotating}
\usepackage{graphics}
\usepackage{booktabs}
\usepackage{tikz}
\usepackage{array}
\usepackage{calc}
\usepackage{hhline}
\usepackage{booktabs}
\usepackage{siunitx}
\usepackage{tabularx}
\usepackage{tablefootnote} 
\usepackage[T1]{fontenc}

\usepackage[utf8]{inputenc}

\usepackage{microtype}

\usepackage{inconsolata}

\def\europa{\mbox{\textsc{Europa}}}

\usepackage{xcolor}
\usepackage{soul}

\newcolumntype{Y}{>{\centering\arraybackslash}X}

\newlength\myfntht

\title{{\LARGE \europa{}}: A Legal Multilingual Keyphrase Generation Dataset}

\author{Olivier Salaün, Frédéric Piedboeuf, Guillaume Le Berre \\ \textbf{David Alfonso Hermelo, Philippe Langlais} \\
{\normalfont RALI, DIRO, Université de Montréal, Canada}
\\
\texttt{\{olivier.salaun, frederic.piedboeuf,}
\\
\texttt{guillaume.le.berre, philippe.langlais\}@umontreal.ca}
\\
\texttt{david.alfonso.hermelo@gmail.com}
}

\begin{document}
\maketitle
\begin{abstract}
Keyphrase generation has primarily been explored within the context of academic research articles, with a particular focus on scientific domains and the English language. 
In this work, we present \europa{}, a dataset for multilingual keyphrase generation in the legal domain.
It is derived from legal judgments from the Court of Justice of the European Union (EU), and contains instances in all 24 EU official languages.
We run multilingual models on our corpus and analyze the results, showing room for improvement on a domain-specific multilingual corpus such as the one we present.
\end{abstract}

\section{Introduction}

Keyphrases are short phrases that describe a text, and have been shown to be useful for many applications, from document indexing~\citep{medelyan2006thesaurus} to opinion mining~\citep{berend-2011-opinion}.
While heavily researched for the STEM (science, technology, engineering, and mathematics) domains, there has been little investigation on its application to law. 
This is surprising, as keywords can reduce the workload of legal experts by allowing them to get the gist of lengthy documents~\cite{mandal2017automatic, sakiyama2023automated, cerat2023lexkey}.

%

The scope of this work is to assess to what extent keyphrases can be automatically generated in the legal domain. 
Our contributions are the following:
(1)~We collected and curated \europa{}, the first open benchmark for legal KPG (keyphrase generation), spanning 24 languages and extracted from real-world European judgments.\footnote{The dataset can be downloaded from the Hugging Face hub at \url{https://huggingface.co/datasets/NCube/europa} while evaluation scripts and model outputs are available at \url{https://github.com/rali-udem/europa}.} (2)~We provide in-depth analysis of our corpus, highlighting its differences compared to other existing corpora.
(3)~We report performances achieved by multilingual generative models on this benchmark and point out areas where performances could be improved.

\section{Related Work}

While English KPG has been studied fairly extensively in technical and academic domains, our work bridges two less common fields~: legal KPG and multilingual KPG. 
In this section, we first review the literature surrounding keyphrase extraction (KPE) and generation before covering legal KPE and KPG. 
For a more complete survey on KPG, we refer to~\citet{xie2023statistical}.

\subsection{Keyphrase Extraction and Generation}

Keyphrases may be either \textbf{present} in the input (could be retrieved with purely \textbf{extractive} methods) or \textbf{absent} from it (needing \textbf{abstractive} methods to generate them). 

Many extractive methods have been suggested and used over the years, either using unsupervised heuristic rules and pre-trained machine learning (ML) models to detect and extract keyphrases~\citep{liu-etal-2010-automatic, gollapalli2014extracting, meng-etal-2017-deep, yu2018wikirank} or using ML/deep learning (DL) classifiers trained on labelled data to learn to identify them ~\citep{yang2018task, chen-etal-2018-keyphrase, wang2018exploiting, basaldella2018bidirectional, alzaidy2019bi, sun2019divgraphpointer, mu2020keyphrase, sahrawat2020keyphrase}. 
Extractive methods are however incomplete as, for most domains, a large portion of the target keyphrases are absent from the document, hence an increased interest in generative models in more recent works.

While research on KPG initially used seq-to-seq models in the form of RNNs~\cite{meng-etal-2017-deep, meng2019does, meng-etal-2021-empirical, yuan2020one}, modern KPG uses mostly fine-tuning of pre-trained transformer models~\citep{vaswani2017attention}, such as BART~\citep{lewis-etal-2020-bart}, which showed itself to be as efficient as previous, more complex RNN models~\cite{chowdhury2022applying}.
Current English state-of-the-art results are currently obtained by using BART models which are further pre-trained for scientific KPG (KeyBART by~\citet{kulkarni-etal-2022-learning} and SciBART by~\citet{wu2022pre}).
KPG has been conducted on limited domains, such as academic and STEM papers~\citep{hulth2003improved, nguyen2007keyphrase, kim-etal-2010-semeval, meng-etal-2017-deep, krapivin2009large} or news~\cite{gallina2019kptimes, koto2022lipkey}, although the need for KPG extends beyond those domains.

Of particular interest is KPG for long documents, which has benefited from a few studies, especially in recent years.~\citet{ahmad-etal-2021-select} proposed a two-step approach of this problem, where they first select salient sentences before using those to generate keyphrases, while~\citet{garg-etal-2022-keyphrase} tested multiple ways of including information from the main body (generated summary, citation sentences, random sentences, etc), showing that adding a generated summary was the most efficient strategy.~\citet{mahata2022ldkp} proposed a corpus for scientific KPG using the whole document instead of only the title and abstract, which is what is currently used in KPG. They do not however run any baseline on that corpus.


\subsection{Keyphrase Generation in the Context of Legal Domain} 


Legal documents are more complex and longer than those commonly used in NLP, as illustrated by the European Court of Human Rights corpus~\citep{chalkidis-etal-2019-neural}.
This motivates a high demand from legal professionals for automatic digests of such documents \citep{bendahman2023quelles}, mostly from the angle of legal summarization.
This task, more widespread than legal KPG, was first formalized as an extractive task in which the most salient segments from the document are returned as a summary~\citep{farzindar-lapalme-2004-legal, farzindar2004letsum, saravanan2010identification, polsley-etal-2016-casesummarizer, aumiller-etal-2022-eur}. 
Similarly to KPG tasks, abstractive summarization models based on transformer architecture~\citep{vaswani2017attention} became more widespread than extractive ones.
However, open legal summarization benchmarks such as BigPatent~\citep{sharma-etal-2019-bigpatent}, Multi-LexSum~\citep{shen2022multilexsum}, and EUR-Lex-Sum~\citep{aumiller-etal-2022-eur}, are still scarce.

Legal KPG may be considered as a specialized type of summarization but, to the best of our knowledge, no open benchmark is available for such a task.
Moreover, most existing works have been focusing on keyphrase extraction~\citep{le2013unsupervised, audich2016extracting, mandal2017automatic, daojian2019keyphrase}, ignoring absent keyphrases.
The very first legal KPG experiments addressing abstractive keyphrases were conducted by~\citet{cerat2023lexkey} and~\citet{sakiyama2023automated}, but within a monolingual setting and with no public dataset release. This is a major issue as generative models are particularly data-greedy.
Furthermore, language-wise, most of the open legal datasets are in English~\citep{chalkidis-etal-2023-lexfiles} and, when it comes to existing open multilingual legal benchmarks \cite{chalkidis-etal-2021-multieurlex, savelka2021lex, aumiller-etal-2022-eur, niklaus2023lextreme}, none are related to KPG.
Overall, our contribution aims at fulfilling these gaps by providing an open multilingual dataset for legal keyphrase generation.


\begin{table*}[!ht]
\centering
{\renewcommand{\arraystretch}{1.5}
\begin{tabularx}{\textwidth}{lX}
\hline
\textbf{\makecell[tl]{Input text\\(fr)}} & La demande de décision préjudicielle porte sur l’interprétation de l’article 48 du règlement (CEE) n° 1408/71 du Conseil,
du 14 juin 1971, relatif à l’application des régimes de sécurité sociale aux travailleurs salariés, aux travailleurs non salariés
et aux membres de leur famille qui se déplacent à l’intérieur de la Communauté (JO L 149, p. 2) [...] \\
\textbf{KPs (fr)} & Assurance vieillesse -- Travailleur ressortissant d’un État membre -- Cotisations sociales -- Périodes différentes -- États membres différents -- Calcul des périodes d’assurance -- Demande de pension -- Résidence dans un État tiers \\
\hline
\textbf{\makecell[tl]{Input text\\(de)}} & Das Vorabentscheidungsersuchen betrifft die Auslegung von Art. 48 der Verordnung (EWG) Nr. 1408/71 des Rates vom 14. Juni 1971 zur Anwendung der Systeme der sozialen Sicherheit auf Arbeitnehmer und Selbständige sowie deren Familienangehörige, die
innerhalb der Gemeinschaft zu‑ und abwandern (ABl. L 149, S. 2) [...] \\
\textbf{KPs (de)} & Altersversicherung -- Arbeitnehmer mit der Staatsangehörigkeit eines Mitgliedstaats -- Sozialbeiträge -- Unterschiedliche Zeiten -- Unterschiedliche Mitgliedstaaten -- Berechnung der Versicherungszeiten -- Rentenantrag -- Wohnort in einem Drittland \\
\hline
\textbf{\makecell[tl]{Input text\\(en)}} & This reference for a preliminary ruling concerns the interpretation of Article 48 of Council Regulation (EEC) No 1408/71 of
14 June 1971 on the application of social security schemes to employed persons, to self-employed persons and to members of
their families moving within the Community (OJ, English Special Edition 1971 (II), p. 416) [...] \\
\textbf{KPs (en)} & Old-age insurance -- Worker who is a national of a Member State -- Social security contributions -- Separate periods -- Different Member States -- Calculation of periods of insurance -- Application for a pension -- Residence in a non-Member State \\
\hline
\end{tabularx}}
\caption{Example of a \href{https://eur-lex.europa.eu/legal-content/EN/TXT/?uri=CELEX:62006CJ0331}{judgment} available in 22 languages with the corresponding input text and target keyphrases (French, German and English pairs are shown above). During the KPG task, keyphrases are separated by semicolons instead of dashes to ensure consistency with other KPG datasets.} 
\label{tab:example}
\end{table*}


\section{The Creation of the \europa{} Dataset}

Cases in the Court of Justice of the European Union (CJEU),  which aims at ensuring the consistent interpretation and application of the EU law across all EU institutions~\citep{cjeu2023annual_report}, can be processed in any of the 24 EU official languages, depending on the Member State involved, making drafting and translation pivotal and complex tasks. Stakeholders and judges communicate in their respective languages and rely on lawyer-linguists for document exchange. 
Once a judgment is rendered, it is translated into other EU languages to ensure consistency in the judgment text, keyphrases, and their legal implications \citep{domingues2017multilingual}. Through private correspondence, the CJEU explained that keyphrases, drafted by the Registry and completed by the reporting judge's cabinet or the advocate general, aim at providing concise case descriptions.
Their meticulous construction establishes them as high-quality gold references for KPG evaluation.

\subsection{Data Collection \& Processing} \label{sec:data_collection_processing}

For the sake of clarity, we define a \textbf{judgment} as a collection of documents which refer to the same ruling provided in up to 24 languages.
Each judgment has a unique CELEX identifier that remains the same for every language version available.
Therefore, all language versions with the same CELEX ID are semantically and legally equivalent, and can be considered as parallel.
Within a judgment, each language version is named an \textbf{instance}, a pair comprising the target keyphrases and the input text from the judgment, with both being in the same language.
For example, the judgment shown in Table~\ref{tab:example} has 22 instances spanning 22 languages. 

\begin{figure}
    \centering
    \definecolor{dblue}{HTML}{1745A2}
    \definecolor{lblue}{HTML}{1E80A4}
    \scalebox{0.80}{\renewcommand{\arraystretch}{1.2}
    \begin{tabular}{p{1.1\linewidth}}
        \textsf{\textbf{\color{dblue}Judgment of the Court (Second Chamber) of 3 April 2008.}} \\
        \textsf{\textbf{\color{dblue}John Doe v Raad van Bestuur van de Sociale Verzekeringsbank.}} \\ 
        \textsf{\textbf{\color{dblue}Reference for a preliminary ruling: Rechtbank te Amsterdam - Netherlands.}} \\
        \textsf{\textbf{\color{dblue}\hl{Old-age insurance - Worker who is a national of a Member State - Social security contributions - Separate periods - Different Member States - Calculation of periods of insurance - Application for a pension - Residence in a non-Member State.}}} \\ 
        \textsf{\textbf{\color{dblue}Case C-331/06.}} \\
        \textsf{\textit{ECLI identifier: ECLI:EU:C:2008:188}} \\
        \textsf{\textbf{\color{lblue}Form:} Judgment} \\
        \textsf{\textbf{\color{lblue}Author:} Court of Justice} \\
        \textsf{\textbf{\color{lblue}Date of document:} 03/04/2008} \\
    \end{tabular}}
    \caption{Query result with multi-line title of a case containing target keyphrases in English (highlighted). The case is the same as the one from Table~\ref{tab:example}. Non-highlighted text is excluded from targets as it has low value from KPG standpoint.}
    \label{fig:snippet}
\end{figure}

In May 2023, we performed a first pass on the EUR-Lex database\footnote{\url{https://eur-lex.europa.eu}}, the main legal EU online database \citep{bernet2006eur}, scraping 304\,426 query results corresponding to 19\,319 judgments released by the CJEU.
These query results (each being a potential instance for our corpus) consist in small snippets (as the one in Figure~\ref{fig:snippet}) containing the multi-line title and meta information of the case including the document identifier, but not the judgment text. Then, we made a second pass in order to scrape the plain text of the judgment using the document identifier collected during the first pass.
The plain text is available as PDF and/or HTML, but we used the latter for convenience.
A total of approximately two weeks was required for collecting the query results snippets and the judgments' HTML files in all 24 languages.
An overview of the corpus collection and curation process is presented in the following paragraphs (more details in Appendix~\ref{app:preprocessing}).

The structure of a case HTML file generally consists of a mix of keyphrases and meta-information at the top of the document followed by paragraphs that will be merged into the input text.
Separating the top of the document from the paragraphs is crucial in order to ensure that the input text is not contaminated by target keyphrases.
However, identifying keyphrases from the judgment plain text was infeasible as they are surrounded by quotation marks, symbols, and HTML tags that vary across languages and time.
Therefore, the keyphrases were obtained from the small snippet multi-line title as the one in Figure~\ref{fig:snippet}.

Still, in such snippet, the keyphrases are mixed with meta information about the case which we need to get rid of as they are redundant and of low value from a KPG standpoint (e.g. sequences such as \textit{``Reference for a preliminary ruling''} followed by the referring court and the country are so frequent that they would bring noise during training).
Using the \mbox{\texttt{BeautifulSoup}} library\footnote{\url{https://pypi.org/project/beautifulsoup4}}, language-specific regular expressions and domain-aware engineered heuristics, we filtered out the sequence of keyphrases that had the highest number of meaningful phrases.
An additional sanity check was also performed to ensure that keyphrases are properly split (e.g. missing whitespace beside a phrase delimiter) and that the number of phrases remains consistent across languages for each judgment.
Our approach thus ensures the quality of the target keyphrases that will be used in the KPG task during training and evaluation.
For the sake of evaluation consistency with the existing KPG literature, keyphrases are lowercased and separated by semicolons.


Another critical part of corpus preparation is the input text from the judgment HTML files.
These raw documents are delicate to process as they begin with a mix of meta-information (e.g. stakeholders identities) and target keyphrases (which must be excluded from input text), followed by numbered paragraphs.
Moreover, since the HTML tags are inconsistent and vary across years and languages, \texttt{BeautifulSoup} cannot be used to extract the input text from the judgment.
Therefore, we manually designed several language-specific regular expressions to reliably split the documents by matching delimiters in all languages (e.g. \textit{``Judgment''}, \textit{``Sentencia''}, \textit{``Urteil''}). 
By doing so, only the paragraphs of the judgment are retained as a single input string, while the section containing superficial meta-information and target keyphrases is taken away, thus preventing any data leakage.
This was confirmed by a manual examination of 100 random instances, equally distributed across the 24 languages.



After removing instances with empty input or target texts, our final corpus is composed of 17\,833 judgments, in 16 languages on average and spanning cases from 1957 to 2023.
This amounts to a total of 284\,957 instances (input/keyphrases pairs). 
Expectedly, these instances are unevenly distributed across all 24 languages, with languages from older EU Member States being more represented in the dataset. 
For instance, French (the most represented language) amounts to 17\,461 instances (6.13\% of all instances) while Croatian and Irish (a significant outlier) contain 5153 (1.81\%) and 92 (0.03\%) instances, respectively.



\begin{table}[h]
\centering
\begin{tabularx}{\columnwidth}{
    X
    >{\hsize=.5\hsize\centering\arraybackslash}X
    >{\hsize=.5\hsize\centering\arraybackslash}X
    >{\hsize=.5\hsize\centering\arraybackslash}X
    >{\hsize=.5\hsize\centering\arraybackslash}X
}
\toprule
\multirow{2}{*}{\textbf{Split Type}} & \multicolumn{2}{c}{\textbf{Present}} & \multicolumn{2}{c}{\textbf{Absent}} \\
 & \textit{F1@5} & \textit{F1@M} & \textit{F1@5} & \textit{F1@M}\\
\midrule
Random & 30.4 & 41.5 & 15.3 & 18.1\\ 
Temporal & 21.7 & 27.4 & 5.6 & 7.0\\ 
\bottomrule
\end{tabularx}
\caption{Performance for mBART50 with chronological and random splits (weighted average scores). }
\label{tab:random_chrono_splits}
\end{table}

\subsection{Chronological Split}

\begin{table*}[!ht]
\centering
\begin{tabular}{lcccccc}
\toprule
\textbf{Benchmark} & \textbf{\makecell{\# inst.}} & \textbf{\makecell{Avg.\\input\\length}} & \textbf{\makecell{Avg.\\num.\\KPs}} & \textbf{\makecell{\%\\absent\\KPs}} & \textbf{Domain} & \textbf{\makecell{\# lang.}} \\ 
\midrule
Inspec~\citep{hulth-2003-improved} & 2000 & 116 & 9.6 & 30.0 & Comp. Sc. & 1 (en) \\ 
Krapivin~\citep{krapivin2009large} & 2304 & \underline{7696} & 5.3 & 23.5 & Comp. Sc. & 1 (en) \\ 
SemEval~\citep{kim-etal-2010-semeval} & 244 & \textbf{7795} & \underline{15.4} & 18.3 & Comp. Sc. & 1 (en) \\ 
KP20k~\citep{meng-etal-2017-deep} & \textbf{554k} & 146 & 5.3 & 48.7 & Comp. Sc. & 1 (en) \\ 
NUS~\citep{nguyen2007keyphrase} & 211 & 6938 & 11.7 & 17.6 & Science & 1 (en) \\ 
pak2018~\citep{campos2020yake} & 50 & 96 & 3.6 & \textbf{84.2} & Science & 1 (pl) \\
110-PT-BN-KP~\citep{marujo2011keyphrase} & 110 & 301 & \textbf{24.4} & 1.3 & News & 1 (pt) \\ 
WikiNews~\citep{bougouin-etal-2013-topicrank} & 100 & 268 & 9.6 & 4.8 & News & 1 (fr) \\
KPTimes~\citep{gallina2019kptimes} & 280k & 774 & 5.0 & \underline{55.0} & News & 1 (en) \\ 
Papyrus~\citep{piedboeuf2022new} & 30k & 307 & 7.2 & 37.2 & Academic & \underline{18} \\ 
\midrule
\europa{} (ours) & \underline{285k} &  5220 & 7.6 & 52.6 & Legal & \textbf{24} \\ 
\bottomrule
\end{tabular}
\caption{Comparative table among different open KPG benchmarks. \textbf{Avg. input length} refers to the average number of tokens split by whitespaces. Top values are in bold font, second top values are underlined.}
\label{tab:benchmarks_comparison}
\end{table*}

It is a common practice to randomly split data in a NLP task \cite{gorman-bedrick-2019-need}. 
However, training and evaluating a model on data from overlapping time periods with similar distributions fails to assess the actual model's temporal generalization \citep{lazaridou2021mind}.
This is why legal NLP generally uses a chronological split of documents instead of combining random shuffle with random split~\citep{chalkidis-etal-2019-neural, medvedeva2021automatic, chalkidis-etal-2021-multieurlex}.
In our case, we tried both splits with a mBART50 model.
On the test set, the performance in terms of $F1@M$ for present keyphrases differs by 14.1 percentage points (11.1 for absent keyphrases) in favour of random split.
Results in Table~\ref{tab:random_chrono_splits} confirm that random split, by ignoring real-world temporal concept drifts, tends to overestimate true performance.
This is consistent with \citet{sogaard-etal-2021-need,  mu-etal-2023-time}.
Therefore, we choose a chronological split for a proper assessment: The training set covers judgments from 1957 to 2010 (131\,076 instances), the validation those from 2011 to 2015 (63\,373 instances), and the test set the ones from 2016 to 2023 (90\,508 instances).\footnote{The temporal split is available at \url{https://huggingface.co/datasets/NCube/europa} and a random split version for those who wish it can be found at \url{https://huggingface.co/datasets/NCube/europa-random-split}}
Full details about documents distribution across these splits are shown in Appendix~\ref{app:stats}.

\section{Dataset Analysis}

As shown in Table~\ref{tab:benchmarks_comparison}, compared to previous works and KPG benchmarks, \europa{} covers more languages and is highly positioned in matters of number of instances, ratio of absent keyphrases, and average input length.

The distribution of keyphrases in our dataset varies across languages due to the fact that the most recent judgments tend to have a higher number of keyphrases attached to them (the average number of keyphrases per language can be found in Appendix~\ref{app:stats}).
Consequently, instances in languages from the most recent Member States tend to be biased towards having more keyphrases.

Another consequence of the temporal evolution of the average number of keyphrases coupled with the chronological split, is that \europa{}'s validation and test sets contain more keyphrases per instance on average (8.3 and 10.5 respectively) compared to its training set (5.4 keyphrases on average).
This creates a discrepancy between the sets that practitioners should be aware of.
However, we advocate that a higher number of keyphrases in the test set of \europa{} will be beneficial to the evaluation of the competing models by providing a larger number of potential gold keyphrases, and by assessing the capacity of models to generalize across time when target keyphrases follow different patterns with respect to the past.

\begin{table*}
\centering
\resizebox{\textwidth}{!}
{
\begin{tabular}{l|ccccccc|ccccccc}
\toprule
\multirow{3}{*}{\textbf{Model}} & \multicolumn{7}{c|}{\textbf{Weighted Average}} & \multicolumn{7}{c}{\textbf{Unweighted Average}} \\
 & \multicolumn{3}{c}{\textbf{F1 Present}} & \multicolumn{3}{c}{\textbf{F1 Absent}} & \textbf{MAP} & \multicolumn{3}{c}{\textbf{F1 Present}} & \multicolumn{3}{c}{\textbf{F1 Absent}} & \textbf{MAP} \\
 & \textit{@5} & \textit{@10} & \textit{@M} & \textit{@5} & \textit{@10} & \textit{@M} & \textit{@50} & \textit{@5} & \textit{@10} & \textit{@M} & \textit{@5} & \textit{@10} & \textit{@M} & \textit{@50} \\
\midrule
YAKE & 2.0 & 2.2 & 2.2 & 0.0 & 0.0 & 0.0 & 0.2 & 1.9 & 2.1 & 2.1 & 0.0 & 0.0 & 0.0 & 0.2\\
mT5-small & 11.6 & 7.7 & 17.4 & 2.6 & 1.7 & 4.1 & 4.3 & 11.1 & 7.4 & 16.6 & 2.5 & 1.6 & 3.9 & 4.1\\
mT5-base & 13.2 & 8.8 & 19.5 & 3.4 & 2.3 & 5.4 & 5.5 & 12.7 & 8.5 & 18.8 & 3.3 & 2.2 & 5.2 & 5.3\\
mT5-large\textsuperscript{a,b} & 13.2 & 8.8 & 19.5 & 3.4 & 2.3 & 5.5 & 5.6 & 12.6 & 8.4 & 18.6 & 3.3 & 2.2 & 5.2 & 5.4\\
mBART50 & \underline{21.7} & \underline{14.6} & \underline{27.4} & \underline{5.6} & \underline{3.8} & \underline{7.0} & \underline{11.4} & \underline{20.8} & \underline{14.0} & \underline{26.3} & \underline{5.3} & \underline{3.6} & \underline{6.7} & \underline{10.9}\\
mBART50-8k\textsuperscript{a} & \textbf{23.9} & \textbf{16.3} & \textbf{29.3} & \textbf{5.8} & \textbf{3.9} & \textbf{7.4} & \textbf{12.3} & \textbf{23.0} & \textbf{15.6} & \textbf{28.2} & \textbf{5.6} & \textbf{3.8} & \textbf{7.1} & \textbf{11.8} \\
\bottomrule
\end{tabular}}
\caption{Weighted and Unweighted average F1@$k$ scores over all languages (\%). MAP refers to MAP@50 for all keyphrases combined. Detailed results per language are shown in Appendix~\ref{app:scores_per_language_for_each_model}. Highest and second highest scores are in bold font and underlined, respectively. Each model is run once. \textsuperscript{a}~Due to greater number of parameters and higher training costs, these models were trained during 5 epochs. \textsuperscript{b}~With a 6\textsuperscript{th} training epoch, mT5-large outperforms mT5-base for some metrics by at most 0.2 percentage points.} 
\label{tab:all_scores_mu}
\end{table*}

\section{Models} 


As the majority of \europa{}'s target keyphrases are absent from input documents, we focus on generative models, as extractive ones such as YAKE \citet{campos2020yake} are ill-suited here.
Recent state-of-the-art models in English KPG rely heavily on pretrained models, such as KeyBART~\citep{kulkarni-etal-2022-learning} or SciBART~\citep{beltagy-etal-2019-scibert}, which are pre-trained and fine-tuned on a massive corpus of English scientific documents.
However, such models are not available in a multilingual format, nor are they specific to the legal domain.
We therefore choose to use mBART50~\citep{tang2020multilingual}, a variant of mBART~\citep{liu-etal-2020-multilingual-denoising} with support for 50 languages instead of 25.
We also test mT5~\citep{xue-etal-2021-mt5}, which covers 101 languages.

Most models, including mBART50, have a maximum input sequence length typically set to 1024 tokens.
While mT5's input length can be set arbitrarily, it was originally pretrained with a context of 1024.
The main caveat is mT5's memory complexity  that increases quadratically with input length, hence a prohibitive computing cost. 
While there exists some models whose maximum input length reaches up to 16k tokens, such as Longformer~\citep{beltagy2020longformer}, BigBird~\citep{zaheer2020big} or LongT5~\citep{guo-etal-2022-longt5}, these remain computationally expensive to run with our current resources and are only suitable for English data. 
Therefore, similarly to \citet{cerat2023lexkey}, we implemented a mBART variant with LSG attention \citep{condevaux2023lsg} such that the maximum input length reaches 8192 tokens instead of 1024. 
Doing so makes the memory complexity increase linearly with respect to the input length, instead of quadratically as it is the case with a traditional attention mechanism.
All models are trained with early stopping and a maximum epoch number of 10, except mBART50-8k and mT5-large with only 5 epochs, as their training is more time-consuming. 
Details about the hyperparameters and training process are provided in Appendices~\ref{app:hyperparameters} and~\ref{app:training_inference_times}.

\subsection{Evaluation Protocol}

As has become common practice in recent works on KPG~\citep{meng-etal-2017-deep, garg-etal-2023-data, shen2023enhanced, chen2023enhancing}, model evaluation relies on two F1 measures: F1@$k$ ($k= \{5, 10 \} $) and F1@$M$, calculated separately for present and absent keyphrases.
F1@$M$ is computed using the entirety of the generated keyphrases while F1@$k$ is computed using the best $k$ generated keyphrases (by truncating if necessary). 
In both cases, the number of target keyphrases remain untouched.
F1@$k$ is calculated using only the top $k$ best scoring keyphrases among the model's predictions, hence an upper boundary below 1 whenever the number of target keyphrases exceeds $k$, which occurs in most of \europa{}'s instances.
This is why F1@$M$ tends to better reflect the model performance as all candidate keyphrases are taken into account without being truncated.
For more details about these metrics, we refer to \citet{yuan-etal-2020-one}.
Following the literature, target and predicted keyphrases are lowercased and stemmed (e.g. \citet{meng-etal-2017-deep} applied Porter Stemmer for English) before conducting an exact match.
Stemming is a critical step without which a candidate keyphrase could be errouneously considered wrong because of the morphological nature of the language.
That is why we applied stemming for all languages for treating them as fairly as possible.
For most of them, the Snowball stemmer \cite{porter2001snowball} was used.
In addition to F1 measures, we also computed MAP@50 (Mean Average Precision).

\begin{table}[t]
\centering
\begin{tabularx}{\columnwidth}{
    X
    >{\hsize=.5\hsize\centering\arraybackslash}X
    >{\hsize=.5\hsize\centering\arraybackslash}X
    >{\hsize=.5\hsize\centering\arraybackslash}X
    >{\hsize=.5\hsize\centering\arraybackslash}X
}
\toprule
\multirow{2}{*}{\textbf{Language}} & \multicolumn{2}{c}{\textbf{Present}} & \multicolumn{2}{c}{\textbf{Absent}} \\
 & 1k & 8k & 1k & 8k \\
\midrule
French & 33.1 & \textbf{34.4} & 8.7 & \textbf{9.6}\\
German & 33.5 & \textbf{35.8} & 4.9 & \textbf{5.0}\\
English & 29.5 & \textbf{31.7} & 5.0 & \textbf{5.1}\\
Italian & 30.5 & \textbf{33.3} & \textbf{3.7} & 3.6\\
Dutch & 31.9 & \textbf{34.4} & 3.5 & \textbf{3.7}\\
Greek & 14.8 & \textbf{15.7} & 10.6 & \textbf{11.1}\\
Danish & 30.4 & \textbf{32.6} & \textbf{3.2} & 2.6\\
Portuguese & 27.6 & \textbf{30.1} & 5.5 & \textbf{6.5}\\
Spanish & 29.2 & \textbf{31.8} & 4.8 & 4.8\\
Swedish & 33.1 & \textbf{33.9} & 4.2 & \textbf{5.0}\\
Finnish & 27.5 & \textbf{28.9} & 11.6 & \textbf{13.1}\\
Lithuanian & \textbf{34.9} & 32.9 & 4.2 & \textbf{4.7}\\
Estonian & 29.5 & \textbf{31.0} & \textbf{4.7} & 4.5\\
Czech & 26.4 & \textbf{29.5} & 13.7 & \textbf{14.6}\\
Hungarian & 14.3 & \textbf{15.3} & 10.4 & \textbf{11.3}\\
Latvian & \textbf{34.7} & 33.1 & 3.5 & \textbf{3.6}\\
Slovene & 25.6 & \textbf{29.2} & 11.1 & \textbf{11.8}\\
Polish & 26.3 & \textbf{29.7} & 11.7 & \textbf{13.2}\\
Maltese & 24.7 & \textbf{27.0} & \textbf{7.7} & 7.1\\
Slovak & \textbf{25.2} & 24.9 & 15.7 & \textbf{16.0}\\
Romanian & 30.7 & \textbf{33.8} & 5.4 & \textbf{5.7}\\
Bulgarian & 29.0 & \textbf{29.2} & 3.4 & \textbf{3.5}\\
Croatian & 8.6 & \textbf{15.5} & \textbf{4.1} & 3.8\\
Irish & 0.0 & \textbf{2.2} & 0.0 & 0.0\\
\bottomrule
\end{tabularx}
\caption{Side-by-side F1@M scores comparison per language between mBART50 and mBART50-8k.}
\label{tab:scores_mbart1k_mbart8k_f1atM}
\end{table}

\begin{table*}[h!]
\centering
\begin{tabular}{p{0.97\textwidth}}
\toprule
\textbf{Reference}: \textcolor{blue}{Appeal} \textcolor{gray}{[5]} -- \textcolor{blue}{Competition} \textcolor{gray}{[5]} -- \textcolor{blue}{\textit{Agreements, decisions and concerted practices}} \textcolor{gray}{[3]} -- Pharmaceutical products -- \textit{Market for antidepressant medicines (citalopram)} -- \textit{Settlement agreements concerning process patents concluded between a manufacturer of originator medicines holding those patents and manufacturers of generic medicines} -- \textcolor{blue}{Article 101 TFEU} \textcolor{gray}{[1]} -- Potential competition -- Restriction by object -- Characterisation -- \textit{Calculation of the amount of the fine} \\
\textbf{mT5-small}: \textcolor{blue}{Appeal} -- \textcolor{blue}{Competition} -- Regulation (EC) No 1/2003 \\
\textbf{mT5-base}: \textcolor{blue}{Appeal} -- \textcolor{blue}{Competition} -- \textcolor{blue}{\textit{Agreements, decisions and concerted practice}} \\
\textbf{mT5-large}: \textcolor{blue}{Appeal} -- \textcolor{blue}{Competition} -- Regulation (EC) No 1/2003\\
\textbf{mBART50}: \textcolor{blue}{Appeal} -- \textcolor{blue}{Competition} -- \textcolor{blue}{\textit{Agreements, decisions and concerted practices}} -- Non-contractual liability of the Community -- \textcolor{purple}{Guidelines on the application of Article 101 TFEU} -- Principle of proportionality -- Obligation to state the reasons on which the decision is based \\
{\textbf{mBART50-8k}: \textcolor{blue}{Appeal} -- \textcolor{blue}{Competition} -- \textcolor{blue}{Article 101 TFEU} -- \textcolor{blue}{\textit{Agreements, decisions and concerted practices}} -- \textcolor{purple}{Antidepressant medicinal products} -- Article 23(2)(a) of Regulation (EC) No 1/2003 -- Article 23(2)(a) of Regulation (EC) No 1/2003 -- \textcolor{purple}{Concept of ‘restrictions of competition by object’} -- \textcolor{purple}{Reduction of the amount of the fine}} \\
\bottomrule
\end{tabular}
\caption{Example of generated keyphrases with an instance in English. \textcolor{blue}{Blue} phrases are exact matched targets, \textcolor{purple}{purple} ones are candidates which could be relevant but are not rewarded by the exact match evaluation approach. We added comments in \textcolor{gray}{grey} with the number of times some targets were matched. Absent keyphrases are in \textit{italics}.} 
\label{tab:generated_kps}
\end{table*}

\section{Results}

$F1@5$, $F1@10$ and $F1@M$ scores for present and absent keyphrases over all languages are reported in Table~\ref{tab:all_scores_mu}.
Since the number of instances varies across languages, we report average scores that are \textbf{weighted} and \textbf{unweighted}.
The former is an average score across all instances without taking language into consideration.
Consequently, it can be highly influenced by high-resource languages covering more instances.
The latter is an unweighted average among languages' scores.
In other words, all languages have equal importance, thus reflecting the ability of a model to perform equally well in high- and low-resource languages.

Unsurprisignly, the YAKE extractive model performs poorly, finding none of the absent keyphrases.
With a fixed input length of 1024 tokens, the three mT5 variants dramatically underperform mBART50.
This is surprising as mT5 covers twice as many languages than mBART50.
Also, as mT5-large has 59\% more parameters than mBART50, it would have been expected to outperform the latter, but the results reveal otherwise.
When comparing mBART50 with mBART50-8k, the increase in the maximum context length brings significant improvement across all metrics.
While an average gain of around 2\% (for $F1@M$ over present phrases) may not seem significant, KPG evaluation often greatly underestimates the true performance of the models, due to the difficulty of correctly evaluating whether a keyphrase is correct or not \citep{wu2023kpeval}. 
As such, the improvement shown by the mBART50-8k model is significant and reflected in the generated keyphrases, thus emphasizing the benefits in enlarging the maximum input length.
Still, Table~\ref{tab:scores_mbart1k_mbart8k_f1atM} shows that input context enlargement does not have uniform effects across languages.
For instance, for present keyphrases, some languages get small gains (Greek, Swedish), and some degrade (Lithuanian, Latvian).
For high-resource languages such as English and Italian, performance improves on present phrases, but seems stagnant on absent ones.
For low-resource languages, Croatian has the most dramatic improvement on present phrases while Irish gets one score above $0$ with mBART50-8k.
However, the performance for these languages still lags behind that of moderate-resource ones.
This is understandable as these languages have almost no training instances in our temporal split setting, thus revealing the difficulty of conducting KPG for unseen languages.\footnote{With a random split dataset, $F1@M$ achieved by mBART50 for Croation/Irish reaches 46.9/20.5 for present keyphrases and 13.9/10.2 for absent ones}

\section{Analysis and Discussion}

The first striking observation is that mBART models, despite covering less languages compared to mT5s, succeed in outperforming the latter models.
One explanation is that mT5 small, base and large variants generated on average significantly fewer phrases per instance (2.1, 2.3 and 2.3, respectively) with respect to mBART50 and mBART50-8k (5.5 and 6.0).
Consequently, mT5 models are less likely to achieve high scores.
The other noticeable result is that the mBART models even succeed in outperforming mT5s in languages that only mT5s support such as Bulgarian or Greek.
This suggests that \citet{conneau-etal-2020-unsupervised}'s corpus used for pretraining mBART50 gave it a capacity to deal with much more languages than stated by \citet{tang2020multilingual}.

The improvement from mBART50 to mBART50-8k is significant and emphasizes the importance of larger input length.
Less that 14\% of all instances fit into a 1024-tokens context window.
Around 58\% do when that length reaches 8k tokens.
Building models that can efficiently generate keyphrases from larger documents is therefore crucial in order to achieve further progress (the length reaches 9160 tokens on average, and 17k at the 90th percentile, with mBART50 tokenizer).\footnote{We tried a sliding-window-based model that was not conclusive, thus the need to find better ways to capture context.}

Moreover, KPG models struggle for keyphrases with more tokens (split by whitespaces).
In the case of mBART50-8k, a matched target keyphrase contains on average 3.1 tokens, while an unmatched target phrase contains 5.7 tokens.
Unmatching generated keyphrases contain 7.7 tokens on average. 
Upon manual inspection of generated keyphrases, most models succeed in generating simple phrases made of up to 3 terms, but they indeed struggle for longer noun phrases that refer to very specific or technical concepts such as \textit{``Market for antidepressant medicines (citalopram)''} in Table~\ref{tab:generated_kps}.
Also, although some candidate keyphrases are still relevant from a reader's standpoint, they are penalized by the exact matching evaluation approach, although stemming is applied.
For instance, the output \textit{``Concept of ‘restrictions of competition by object’"} could be a decent generation for the target \textit{``Restriction by object''} despite the lack of matching.

In order to mitigate this issue, we evaluated the models again with a semantic matching metric \citep{wu2023kpeval} allowing us to compare predictions and targets without having to apply any sort of post-processing or stemming (the tool is however only available for English).
Results in Table~\ref{tab:semantic_matching} have a correlation of $0.99$ with F1@$M$ scores obtained for English for both present and absent keyphrases.
This seems consistent with the ranking among models observed previously in Table~\ref{tab:all_scores_mu}: mBART50-based models outperform mT5-based ones, and mBART50-8k is the front-runner.

\begin{table}
\centering

\begin{tabularx}{\columnwidth}{X
    >{\hsize=.5\hsize\centering\arraybackslash}X
    >{\hsize=.5\hsize\centering\arraybackslash}X
}
\toprule
\textbf{Model} & \textbf{Present} & \textbf{Absent} \\
\midrule
YAKE & 0.335 & 0.155\\
mT5-small & 0.399 & 0.268 \\
mT5-base & 0.422 & 0.283 \\
mT5-large & 0.436 & 0.270 \\
mBART50 & \underline{0.519} & \underline{0.442} \\
mBART50-8k & \textbf{0.548} & \textbf{0.479} \\
\bottomrule
\end{tabularx}
\caption{Semantic matching scores for present and absent keyphrases in English.}
\label{tab:semantic_matching}
\vspace{-10px}
\end{table}


\section{Conclusion}

In this paper, we present \europa{}, a novel and open multilingual keyphrase generation dataset in the legal domain. 
We believe this dataset can help alleviate two current shortcomings of the keyphrase generation task: the lack of data in domains other than STEM; and the lack of multilingual non-English datasets.
Our dataset is available at \url{https://huggingface.co/datasets/NCube/europa}.
We furthermore provide an analysis of \europa{} with key statistics, thus giving insights on the particularities of our dataset. Finally, we run multiple models in various settings in order to give an initial point of comparison for future works on \europa{}.
Our corpus also highlights the need to efficiently capture larger input context, and will be a suitable testbed for models designed to do so.

\section*{Limitations}

\textbf{Low Resource Languages:} The choice of a chronological split, though enabling a realistic KPG performance assessment with quasi-equal amount of test instances for each language (except Irish), results in dramatic differences in available training data across languages.
This is particularly true for the latest official EU languages, such as Croatian and Irish.
One possible approach for mitigating this issue in future works would be to complement the training data with other legal documents released by other EU institutions.
A random split version of our dataset is available for those who would like an identical distribution of languages across training, validation and test sets (\url{https://huggingface.co/datasets/NCube/europa-random-split}).



\textbf{Computing Cost Scalability:}
The architecture of most of the models used here is derived from experiments in which the text input length rarely exceeds 1024.
This is a major caveat when deploying such models on real-world data with significantly longer documents, such as the legal documents in our dataset which often exceed several thousand tokens.
Moreover, dealing with documents whose language is not supported by the tokenizer mechanically increases the number of tokens.
This is due to the fact that running and training transformer models with higher maximum input sequences is costly both in terms of time and GPU memory.
The computing burden was already emphasized by \citet{aumiller-etal-2022-eur} of legal summarization for EU documents and \citet{sakiyama2023automated} for legal KPG in Portuguese.


\section*{Ethics Statement}

\textbf{Copyright:} The Publications Office of the European Union gave us the written confirmation that cases of the Court of Justice of the European Union could be used for commercial and non-commercial purposes, as stated in the European Commission decision released on 12 December 2011~\citep{commission2011reuse}. 
Our models are implemented with \citet{wolf-etal-2020-transformers}'s \texttt{transformers} library licensed under Apache 2.0 while evaluation is performed with \citet{meng-etal-2017-deep}'s tools available under the MIT license. 
Therefore, one is granted permission to modify and distribute the licensed material.

\textbf{Personal Data Protection:}
According to the CJEU personal \href{https://curia.europa.eu/jcms/upload/docs/application/pdf/2015-11/anonymat_notice_cj.pdf}{data protection policy}, the Court anonymizes personal information upon request of a party, referring court/tribunal or upon its own volition.
In that case, anonymity is granted throughout the entire procedure according to article 95 of \href{https://eur-lex.europa.eu/legal-content/EN/TXT/?uri=celex:32012Q0929(01)}{Rules of Procedure of the Court of Justice}.
It must be emphasized that anonymity must be requested at the earliest stage of the proceedings.
Once the decision is drafted and released, there is no a posteriori opt-out option due to the transparency principle with which the CJEU must comply.
Moreover, we have a written confirmation from the Publications Office of the EU that we are allowed to share and redistribute our corpus made from documents publicly available on EU platforms.


\section*{Acknowledgements}

We would like to thank the European Union for granting us access to EUR-Lex data (\url{http://eur-lex.europa.eu}, © European Union, 1998-2023).
We are also grateful to Benjamin Cérat and Di Wu for their advice on some coding implementations.


\bibliography{anthology,custom}
\bibliographystyle{acl_natbib}

\appendix

\section{Hyperparameters}
\label{app:hyperparameters}

While the batch size was set at 16 (with the help of gradient batch accumulation when needed), a grid search was performed for the learning rate with the following values: \num{1e-6}, \num{5e-6}, \num{1e-5}, \num{5e-5}, \num{1e-4}.
For both mBART50 models, we settled on a learning rate of \num{1e-5} which we coupled with the \texttt{AdamW}~\citep{loshchilov2018decoupled} optimizer of \texttt{Pytorch}.
The same applies to mT5 variants but with a learning rate of \num{1e-4}.
All other optimizer hyperparameters are left at default values.
Once the learning rate was determined, we also added 2500 warming steps and a scheduler with a linear decay.
The objective function was cross-entropy loss.
For mBART models, we applied mixed precision (FP16) while brainfloat16 (BF16) was used for mT5 models.
The maximum number of epochs was set to 10 and the model achieving the lowest loss on the validation set is used for the test set.
Patience was set to 5 epochs. 
In the case of large models (mBART-8k, mT5-large), the maximum number of epochs was set to 5 due to much longer training epochs.
For generation, we used a beam search of 10 and return the best sequence, with a maximum output length of 256 tokens.
The hardware consisted in a RTX4090 and a couple of A5000 graphical cards, each with approximately 24 GB of VRAM.
Inference only required a single card for all models, including those with the highest number of parameters.
Overall, the training for either mBART-8k or mT5-large was performed on 2 GPUs for up to 62 hours overall.
For other models, the training lasted for up to 28h on a single GPU (more details in Table~\ref{tab:training_inference_times}). 
The inference on the test set took up to 34 hours, depending on the model.
Given the delay in obtaining keyphrases during inference, the results are based on a single run.

\section{Languages Support per Model}
\label{app:lang_supp}

mBART50 by~\citet{tang2020multilingual} and mT5 by~\citet{xue-etal-2021-mt5}, though presented as multilingual models, do not cover the same languages and both support only partially the official EU languages, as shown in Table~\ref{tab:lang_coverage}.
Although mT5 (its tokenizer remains the same for all variant sizes) supports twice as many languages than mBART, the vocabulary of both tokenizers is roughly of the same size: 250\,112 and 250\,054 tokens, respectively.
Such large vocabulary size is a computing burden during inference time, especially during decoding.

By design, mBART50 tokenizer adds to input and output sequences a token specifying source and target languages (e.g. \texttt{{[en\_XX]}} for English).
As not all EU languages are not covered by mBART50, we manually added to the tokenizer special tokens for all unsupported languages (e.g. \texttt{{[bg\_BG]}} for Bulgarian).
Following mBART50 tokenization process, a language token is added in both input and output sequences.
Unlike translation tasks, our multilingual KPG task is such that, for each instance, the input document and the target keyphrases are both in the same language.
Therefore, the language token remains the same in the input and output within the same instance.

When it comes to mT5, such languages prefixes are not required.
However, applying mBART50 language tokens to mT5-small brought around 1 percentage point improvement across F1@$k$ metrics, thus motivating us to deploy mBART50 language prefixes to the rest of mT5 models.

\begin{table}[t]
    \begin{tabularx}{\columnwidth}{XYY}
    \toprule
        \textbf{Model} & mBART50 & mT5 \\
        \midrule
        \textbf{Total lang.} & 52 & 101 \\
        \midrule
        \textbf{Unsupported EU lang.} & bg, da, el, ga, hu, mt, sk & hr\\
    \bottomrule
    \end{tabularx}
\caption{Language Coverage by Multilingual Models}
\label{tab:lang_coverage}
\end{table}

\newpage

\section{Number of Parameters per Model}


\null
\begin{table}[!h]
\centering
\begin{tblr}{
  column{2} = {r, m},
  cell{1}{2} = {c},
  hline{1,7} = {-}{0.08em},
  hline{2} = {-}{0.05em},
}
\textbf{Model} & {\textbf{\# param.}\\\textbf{(millions)}}\\
google/mt5-small & 172\\
google/mt5-base & 390\\
google/mt5-large & 973 \\
facebook/mbart-large-50 & 611\\
mBART-8k & 626 \\
\end{tblr}
\caption{Number of parameters (in millions) for each model. Each model name corresponds to its identifier on \citet{wolf-etal-2020-transformers}'s Hugging Face platform, expect mBART-8k which we produced by applying LSG attention~\cite{condevaux2023lsg} to mBART.}
\label{tab:nbr_params_per_model}
\end{table}

\section{Training and Inference Times}
\label{app:training_inference_times}

\begin{table}[h!]
\centering
\resizebox{\linewidth}{!}{
\begin{tblr}{
  column{even} = {c},
  column{3} = {c},
  hline{1,7} = {-}{0.08em},
  hline{2} = {-}{0.05em},
}
\makecell[b]{\textbf{Model}} & \makecell{\textbf{Num. of}\\\textbf{training }\\\textbf{epochs}} & \makecell{\textbf{Total }\\\textbf{training}\\\textbf{time}} & \makecell{\textbf{Total }\\\textbf{inference}\\\textbf{time}}\\
mT5-small & 10 & 11h & 7h \\
mT5-base & 10 & 28h & 9h \\
mT5-large* & 5 & 53h & 16h \\
mBART50 & 10 & 25h & 31h \\
mBART50-8k* & 5 & 61h & 34h
\end{tblr}
}
\caption{Total training and inference times for each model. Total training time corresponds to the duration of all training epochs, including computation of loss on the validation set. Total inference time refers to the duration required for generating keyphrases over the entire test set. All measurements correspond to a single run. *Due to greater number of parameters and higher training costs, these models were trained during 5 epochs.}
\label{tab:training_inference_times}
\end{table}

\onecolumn

\null\vfill
\section{Details About the Data Preprocessing}
\label{app:preprocessing}

\null\vfill
\begin{figure*}[h!]
    \centering
    \includegraphics[width=\linewidth]{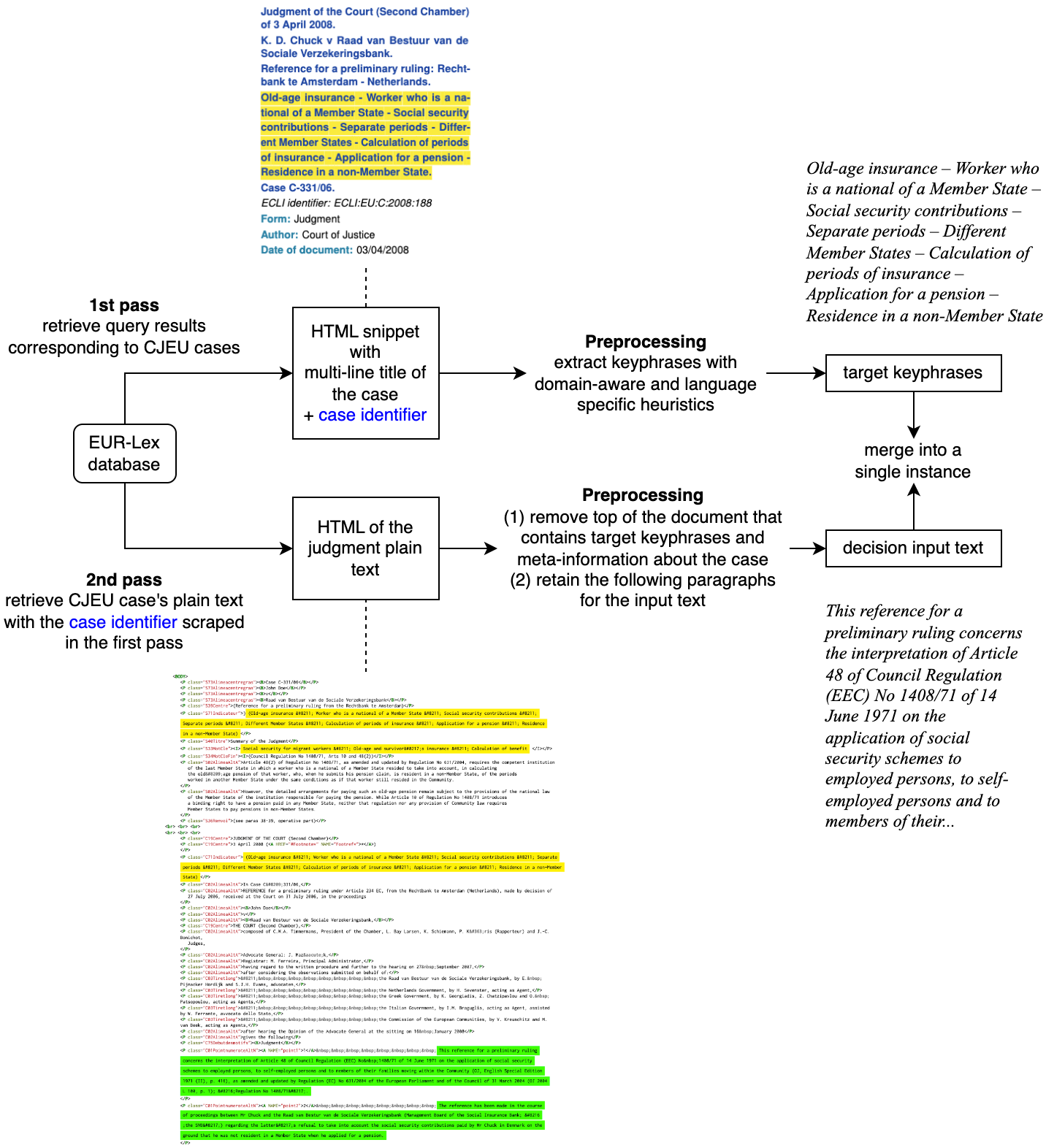}
    \vspace{20px}
    \caption{Diagram illustrating the data collection procedure. Target keyphrases and input text are extracted from separate pieces of HTML that are linked by a common case identifier. After preprocessing, they are merged into a single instance. The same process is repeated in every language.}
    \label{fig:data_collection_process}
\end{figure*}
\vfill

\clearpage

\null\vfill
\section{Additional Statistics}
\label{app:stats}

\begin{table*}[!h]
\centering
\begin{tabular}{lrrccc}
\hline
\textbf{Set} & \textbf{\# judg.} & \textbf{\# inst.} & \textbf{Avg. num. KPs} & \textbf{\% present KPs} & \textbf{\% absent KPs}\\
\hline
Train (1957-2010) & 10\,003 & 131\,076 & 5.4 & 45.6 & 54.4 \\ 
Valid. (2011-2015) & 3\,391 & 63\,373 & 8.3 & 45.1 & 54.9 \\
Test (2016-2023) & 4\,439 & 90\,508 & 10.5 & 51.7 & 48.3 \\
\hline
\end{tabular}
\caption{Statistics on the training, validation, and test sets of \europa{}. \textbf{\# judg.} and \textbf{\# inst.} stand for the numbers of judgments (judgment is a ruling released in several languages) and instances (an instance is a pair with input text and target keyphrases both in a single language).}
\label{tab:stats_sets}
\end{table*}

\begin{table*}[h!]
\centering
\begin{tabular}{lccrrrcccr}
\hline
\textbf{\makecell[l]{Language}} & \textbf{\makecell{Lang.\\ISO\\code}} & \textbf{\makecell{Official\\ EU lang.\\since}} & \textbf{\makecell{Train\\set\\size}}\hspace{-5px} & \textbf{\makecell{Valid.\\set\\size}}\hspace{-5px} & \textbf{\makecell{Test\\set\\size}}\hspace{-5px} & \textbf{\makecell{Avg.\\num.\\KPs}} & \textbf{\makecell{\%\\absent\\KPs}} & \textbf{\makecell{Avg.\\KPs\\length}} & \textbf{\makecell{Avg.\\input\\length}}\hspace{-5px} \\
\hline
French & fr & 1958 & 9692 & 3338 & 4431 & 7.0 & 51.6 & 5.5 & 5651 \\ 
German & de & 1958 & 9028 & 2817 & 3962 & 6.9 & 47.8 & 4.6 & 4877 \\ 
English & en & 1973 & 9047 & 2826 & 3950 & 6.8 & 46.8 & 5.6 & 5539 \\ 
Italian & it & 1958 & 8978 & 2831 & 3946 & 6.9 & 53.2 & 5.6 & 5402 \\ 
Dutch & nl & 1958 & 9066 & 2809 & 3875 & 6.8 & 50.5 & 5.0 & 5374 \\ 
Greek & el & 1981 & 8945 & 2833 & 3917 & 6.9 & 73.9 & 5.3 & 3894 \\ 
Danish & da & 1973 & 9018 & 2702 & 3886 & 6.8 & 52.2 & 4.8 & 5039 \\ 
Portuguese & pt & 1986 & 8347 & 2801 & 3913 & 7.1 & 56.0 & 5.8 & 5763 \\ 
Spanish & es & 1986 & 8462 & 2807 & 3932 & 7.0 & 49.4 & 6.1 & 6154 \\ 
Swedish & sv & 1995 & 6736 & 2788 & 3903 & 7.5 & 45.8 & 4.8 & 5392 \\ 
Finnish & fi & 1995 & 6874 & 2806 & 3886 & 7.5 & 56.5 & 4.0 & 4139 \\ 
Lithuanian & lt & 2004 & 3598 & 2828 & 3908 & 8.5 & 47.1 & 4.8 & 4674 \\ 
Estonian & et & 2004 & 3621 & 2810 & 3912 & 8.4 & 42.4 & 4.0 & 4363 \\ 
Czech & cs & 2004 & 3621 & 2804 & 3913 & 8.4 & 54.9 & 5.1 & 5167 \\ 
Hungarian & hu & 2004 & 3625 & 2805 & 3912 & 8.4 & 55.1 & 4.8 & 5138 \\ 
Latvian & lv & 2004 & 3627 & 2814 & 3896 & 8.5 & 47.1 & 4.8 & 4825 \\ 
Slovene & sl & 2004 & 3568 & 2800 & 3863 & 8.4 & 48.7 & 4.8 & 5187 \\ 
Polish & pl & 2004 & 3526 & 2732 & 3918 & 8.4 & 63.8 & 5.1 & 5385 \\ 
Maltese & mt & 2004 & 3394 & 2739 & 3905 & 8.5 & 58.1 & 5.1 & 5185 \\ 
Slovak & sk & 2004 & 3337 & 2803 & 3894 & 8.5 & 56.9 & 5.1 & 5233 \\ 
Romanian & ro & 2007 & 2484 & 2812 & 3916 & 8.7 & 51.5 & 5.8 & 6085 \\ 
Bulgarian & bg & 2007 & 2480 & 2822 & 3873 & 8.7 & 47.0 & 5.8 & 5827 \\ 
Croatian & hr & 2013 & 2 & 1246 & 3905 & 10.1 & 46.0 & 5.2 & 6367 \\ 
Irish & ga & \href{https://commission.europa.eu/news/irish-now-same-level-other-official-eu-languages-2022-01-03_en}{2022} & 0 & 0 & 92 & 11.9 & 48.6 & 6.5 & 10101 \\ 
\hline
Overall &  & & 131\,076 & 63\,373 & 90\,508 & 7.6 & 52.6 & 5.1 & 5220 \\ 
\hline
\end{tabular}
\caption{Distribution of documents across languages and splits. High-volume languages are those spoken by the earliest EU Member States (e.g. France and Germany are among the Founding States of the EU). \textbf{Avg. KPs length} refers to the average number of words per keyphrase (split at whitespaces). \textbf{Avg. input length} refers to the average number of tokens in the input text (split at whitespaces).} 
\label{tab:distrib_langs}
\end{table*}

\vfill

\clearpage

\null\vfill
\section{Scores per Language for each Model}
\label{app:scores_per_language_for_each_model}

\begin{table*}[h!]
    \centering
    \scalebox{0.9}{
    \begin{tabular}{lcccccccccc}
        \hline
        \multirow{2}{*}{\textbf{Model}} & \multicolumn{2}{c}{\textbf{Bulgarian\,*}} & \multicolumn{2}{c}{\textbf{Croatian\,\textsuperscript{\textdagger}}} & \multicolumn{2}{c}{\textbf{Czech}} & \multicolumn{2}{c}{\textbf{Danish\,*}} & \multicolumn{2}{c}{\textbf{Dutch}} \\
         & \textit{F1@5} & \textit{F1@M} & \textit{F1@5} & \textit{F1@M} & \textit{F1@5} & \textit{F1@M} & \textit{F1@5} & \textit{F1@M} & \textit{F1@5} & \textit{F1@M} \\
        \hline
        YAKE & 1.8 & 1.9 & 2.6 & 2.4 & 3.0 & 2.8 & 0.6 & 1.0 & 2.1 & 2.5 \\ 
        mT5-small & 13.3 & 19.9 & 1.5 & 2.4 & 9.5 & 15.3 & 15.1 & 21.5 & 15.4 & 22.4\\ 
        mT5-base & 14.7 & 21.5 & 3.9 & 6.1 & 10.9 & 17.2 & 16.8 & 23.9 & 16.3 & 23.5 \\ 
        mT5-large & 14.6 & 21.6 & 3.3 & 5.1 & 11.2 & 17.8 & 16.6 & 23.4 & 16.4 & 23.4\\ 
        mBART50 & \underline{23.0} & \underline{29.0} & \underline{5.8} & \underline{8.6} & \underline{20.0} & \underline{26.4} & \underline{25.0} & \underline{30.4} & \underline{25.5} & \underline{31.9} \\ 
        mBART50-8k & \textbf{23.7} & \textbf{29.2} & \textbf{11.8} & \textbf{15.5} & \textbf{23.2} & \textbf{29.5} & \textbf{27.6} & \textbf{32.6} & \textbf{28.3} & \textbf{34.4} \\ 
        
        \hline
    \end{tabular}}

    \vspace{8px}
    \scalebox{0.9}{
    \begin{tabular}{lcccccccccc}
        \hline
        \multirow{2}{*}{\textbf{Model}} & \multicolumn{2}{c}{\textbf{English}} & \multicolumn{2}{c}{\textbf{Estonian}} & \multicolumn{2}{c}{\textbf{Finnish}} & \multicolumn{2}{c}{\textbf{French}} & \multicolumn{2}{c}{\textbf{German}} \\
         & \textit{F1@5} & \textit{F1@M} & \textit{F1@5} & \textit{F1@M} & \textit{F1@5} & \textit{F1@M} & \textit{F1@5} & \textit{F1@M} & \textit{F1@5} & \textit{F1@M} \\
        \hline
        YAKE & 2.7 & 3.4 & 1.9 & 2.5 & 1.1 & 1.2 & 2.4 & 2.5 & 2.4 & 3.1 \\ 
        mT5-small & 15.0 & 20.5 & 13.2 & 18.4 & 9.3 & 14.8 & 15.4 & 23.0 & 16.7 & 23.8 \\ 
        mT5-base & 16.1 & 21.9 & 15.5 & 21.3 & 12.2 & 19.1 & 16.6 & 24.3 & 17.8 & 24.9 \\ 
        mT5-large & 16.4 & 22.2 & 15.6 & 21.5 & 12.0 & 19.0 & 16.7 & 24.5 & 17.9 & 25.2 \\ 
        mBART50 & \underline{24.9} & \underline{29.5} & \underline{25.5} & \underline{29.5} & \underline{19.1} & \underline{27.5} & \underline{26.6} & \underline{33.1} & \underline{28.7} & \underline{33.5} \\ 
        mBART50-8k & \textbf{27.8} & \textbf{31.7} & \textbf{27.2} & \textbf{31.0} & \textbf{21.1} & \textbf{28.9} & \textbf{28.8} & \textbf{34.4} & \textbf{31.1} & \textbf{35.8} \\ 
        \hline
    \end{tabular}}

    \vspace{8px}
    \scalebox{0.9}{
    \begin{tabular}{lcccccccccc}
        \hline
        \multirow{2}{*}{\textbf{Model}} & \multicolumn{2}{c}{\textbf{Greek\,*}} & \multicolumn{2}{c}{\textbf{Hungarian\,*}} & \multicolumn{2}{c}{\textbf{Irish\,*}} & \multicolumn{2}{c}{\textbf{Italian}} & \multicolumn{2}{c}{\textbf{Latvian}} \\
         & \textit{F1@5} & \textit{F1@M} & \textit{F1@5} & \textit{F1@M} & \textit{F1@5} & \textit{F1@M} & \textit{F1@5} & \textit{F1@M} & \textit{F1@5} & \textit{F1@M} \\
        \hline
        YAKE & 1.0 & 0.9 & 2.1 & 1.9 & 0.7 & 1.0 & 2.9 & 3.0 & 2.4 & 2.8 \\ 
        mT5-small & 6.3 & 11.0 & 7.4 & 11.3 & 0.2 & 0.2 & 13.8 & 21.0 & 12.7 & 18.7 \\ 
        mT5-base & 8.3 & 14.2 & 9.1 & 13.7 & \underline{1.3} & \underline{1.8} & 14.8 & 21.9 & 15.0 & 21.4 \\ 
        mT5-large & 7.7 & 13.2 & 9.0 & 13.6 & 0.0 & 0.0 & 14.1 & 20.9 & 14.0 & 20.2 \\ 
        mBART50 & \underline{9.3} & \underline{14.8} & \underline{10.0} & \underline{14.3} & 0.0 & 0.0 & \underline{23.8} & \underline{30.5} & \underline{28.4} & \underline{34.7} \\ 
        mBART50-8k & \textbf{10.7} & \textbf{15.7} & \textbf{10.9} & \textbf{15.3} & \textbf{1.4} & \textbf{2.2} & \textbf{27.2} & \textbf{33.3} & \textbf{27.8} & \textbf{33.1} \\ 
        \hline
    \end{tabular}}

    \vspace{8px}
    \scalebox{0.9}{
    \begin{tabular}{lcccccccccc}
        \hline
        \multirow{2}{*}{\textbf{Model}} & \multicolumn{2}{c}{\textbf{Lithuanian}} & \multicolumn{2}{c}{\textbf{Maltese\,*}} & \multicolumn{2}{c}{\textbf{Polish}} & \multicolumn{2}{c}{\textbf{Portuguese}} & \multicolumn{2}{c}{\textbf{Romanian}} \\
         & \textit{F1@5} & \textit{F1@M} & \textit{F1@5} & \textit{F1@M} & \textit{F1@5} & \textit{F1@M} & \textit{F1@5} & \textit{F1@M} & \textit{F1@5} & \textit{F1@M} \\
        \hline
        YAKE & 1.6 & 1.7 & 0.3 & 0.4 & 2.3 & 2.2 & 2.5 & 2.6 & 1.7 & 1.7 \\ 
        mT5-small & 12.8 & 18.7 & 10.1 & 15.8 & 8.7 & 14.3 & 7.2 & 11.4 & 13.7 & 20.2 \\ 
        mT5-base & 16.2 & 23.3 & 11.3 & 17.7 & 10.7 & 17.6 & 8.4 & 13.0 & 14.0 & 20.5 \\ 
        mT5-large & 15.2 & 21.8 & 11.4 & 17.7 & 10.9 & 17.9 & 8.5 & 13.2 & 14.7 & 21.6 \\ 
        mBART50 & \textbf{28.8} & \textbf{34.9} & \underline{18.8} & \underline{24.7} & \underline{19.1} & \underline{26.3} & \underline{21.6} & \underline{27.6} & \underline{24.9} & \underline{30.7} \\ 
        mBART50-8k & \underline{27.9} & \underline{32.9} & \textbf{21.5} & \textbf{27.0} & \textbf{21.6} & \textbf{29.7} & \textbf{24.6} & \textbf{30.1} & \textbf{28.2} & \textbf{33.8} \\ 
        \hline
    \end{tabular}}

    \vspace{8px}
    \scalebox{0.9}{
    \begin{tabular}{lcccccccc}
        \hline
        \multirow{2}{*}{\textbf{Model}} & \multicolumn{2}{c}{\textbf{Slovak\,*}}& \multicolumn{2}{c}{\textbf{Slovene}} & \multicolumn{2}{c}{\textbf{Spanish}} & \multicolumn{2}{c}{\textbf{Swedish}} \\
         & \textit{F1@5} & \textit{F1@M} & \textit{F1@5} & \textit{F1@M} & \textit{F1@5} & \textit{F1@M} & \textit{F1@5} & \textit{F1@M} \\
        \hline
        YAKE & 1.6 & 1.7 & 2.3 & 2.6 & 2.3 & 2.5 & 2.0 & 2.9 \\ 
        mT5-small & 10.7 & 17.2 & 9.9 & 14.7 & 14.3 & 20.9 & 15.3 & 21.4 \\ 
        mT5-base & 11.4 & 18.2 & 12.1 & 17.6 & 14.9 & 21.8 & 17.2 & 24.2 \\ 
        mT5-large & 12.0 & 19.1 & 12.7 & 18.3 & 15.4 & 22.6 & 16.9 & 23.7 \\ 
        mBART50 & \underline{17.8} & \textbf{25.2} & \underline{20.4} & \underline{25.6} & \underline{23.7} & \underline{29.2} & \underline{28.2} & \underline{33.1} \\ 
        mBART50-8k & \textbf{18.8} & \underline{24.9} & \textbf{23.9} & \textbf{29.2} & \textbf{26.4} & \textbf{31.8} & \textbf{29.3} & \textbf{33.9} \\ 
        \hline
    \end{tabular}}
    \caption{\textbf{Present} keyphrases prediction results for the multilingual setting. Languages with a star (\textasteriskcentered{}) are unsupported by mBART50. Croatian with a dagger (\textdagger) is unsupported by mT5.}
    \label{tab:detailed_scores_present_kps}
\end{table*}
\vfill

\begin{table*}[h!]
    \centering
    \scalebox{0.9}{
    \begin{tabular}{lcccccccccc}
        \hline
        \multirow{2}{*}{\textbf{Model}} & \multicolumn{2}{c}{\textbf{Bulgarian\,*}} & \multicolumn{2}{c}{\textbf{Croatian\,\textsuperscript{\textdagger}}} & \multicolumn{2}{c}{\textbf{Czech}} & \multicolumn{2}{c}{\textbf{Danish\,*}} & \multicolumn{2}{c}{\textbf{Dutch}} \\
         & \textit{F1@5} & \textit{F1@M} & \textit{F1@5} & \textit{F1@M} & \textit{F1@5} & \textit{F1@M} & \textit{F1@5} & \textit{F1@M} & \textit{F1@5} & \textit{F1@M} \\
        \hline
        YAKE & 0.0 & 0.0 & 0.0 & 0.0 & 0.0 & 0.0 & 0.0 & 0.0 & 0.0 & 0.0 \\ 
        mT5-small & 0.9 & 1.6 & 1.2 & 1.7 & 6.0 & 9.5 & 0.8 & 1.3 & 0.9 & 1.4 \\ 
        mT5-base & 1.6 & 2.8 & 2.4 & 3.7 & 6.1 & 9.6 & 1.4 & 2.2 & 1.7 & 2.8 \\ 
        mT5-large & 1.7 & 2.9 & 1.7 & 2.6 & 6.5 & 10.3 & 1.2 & 2.0 & 1.8 & 3.0 \\ 
        mBART50 & \textbf{2.6} & \underline{3.4 }& \textbf{3.6} & \textbf{4.1} & \underline{11.0} & \underline{13.7} & \textbf{2.5} & \textbf{3.2} & \underline{2.8} & \underline{3.5} \\ 
        mBART50-8k & \underline{2.5} & \textbf{3.5} & \underline{3.2} & \underline{3.8} & \textbf{11.7} & \textbf{14.6} & \underline{1.9} & \underline{2.6} & \textbf{2.9} & \textbf{3.7} \\
        \hline
    \end{tabular}}

    \vspace{8px}
    \scalebox{0.9}{
    \begin{tabular}{lcccccccccc}
        \hline
        \multirow{2}{*}{\textbf{Model}} & \multicolumn{2}{c}{\textbf{English}} & \multicolumn{2}{c}{\textbf{Estonian}} & \multicolumn{2}{c}{\textbf{Finnish}} & \multicolumn{2}{c}{\textbf{French}} & \multicolumn{2}{c}{\textbf{German}} \\
         & \textit{F1@5} & \textit{F1@M} & \textit{F1@5} & \textit{F1@M} & \textit{F1@5} & \textit{F1@M} & \textit{F1@5} & \textit{F1@M} & \textit{F1@5} & \textit{F1@M} \\
        \hline
        YAKE & 0.0 & 0.0 & 0.0 & 0.0 & 0.0 & 0.0 & 0.0 & 0.0 & 0.0 & 0.0 \\ 
        mT5-small & 1.0 & 1.8 & 1.0 & 2.0 & 4.3 & 6.5 & 2.6 & 4.5 & 1.2 & 2.0\\ 
        mT5-base & 2.5 & 4.2 & 1.8 & 3.5 & 5.6 & 8.2 & 3.0 & 5.1 & 2.3 & 3.9 \\ 
        mT5-large & 2.3 & 4.0 & 1.9 & 3.6 & 5.4 & 7.9 & 3.0 & 5.0 & 2.3 & 3.9 \\ 
        mBART50 &  \underline{3.7} & \underline{5.0} & \textbf{3.0} & \textbf{4.7} & \underline{10.2} & \underline{11.6} & \underline{7.0} & \underline{8.7} & \textbf{3.6} & \underline{4.9} \\ 
        mBART50-8k & \textbf{3.8} & \textbf{5.1} & \textbf{3.0} & \underline{4.5} & \textbf{11.4} & \textbf{13.1} & \textbf{7.7} & \textbf{9.6} & \textbf{3.6} & \textbf{5.0} \\ 
        \hline
    \end{tabular}}

    \vspace{8px}
    \scalebox{0.9}{
    \begin{tabular}{lcccccccccc}
        \hline
        \multirow{2}{*}{\textbf{Model}} & \multicolumn{2}{c}{\textbf{Greek\,*}} & \multicolumn{2}{c}{\textbf{Hungarian\,*}} & \multicolumn{2}{c}{\textbf{Irish\,*}} & \multicolumn{2}{c}{\textbf{Italian}} & \multicolumn{2}{c}{\textbf{Latvian}} \\
         & \textit{F1@5} & \textit{F1@M} & \textit{F1@5} & \textit{F1@M} & \textit{F1@5} & \textit{F1@M} & \textit{F1@5} & \textit{F1@M} & \textit{F1@5} & \textit{F1@M} \\
        \hline
        YAKE & 0.0 & 0.0 & 0.0 & 0.0 & 0.0 & 0.0 & 0.0 & 0.0 & 0.0 & 0.0 \\ 
        mT5-small & 6.2 & 9.0 & 6.0 & 9.2 & 0.0 & 0.0 & 0.9 & 1.6 & 0.6 & 1.1\\ 
        mT5-base & 7.0 & 10.2 & 6.0 & 9.5 & 0.0 & 0.0 & 1.6 & 2.8 & 1.6 & 2.6 \\ 
        mT5-large & 6.9 & 10.0 & 6.5 & 10.3 & 0.0 & 0.0 & 1.6 & 2.7 & 1.5 & 2.5 \\ 
        mBART50 & \underline{8.7} & \underline{10.6} & \underline{8.0} & \underline{10.4} & 0.0 & 0.0 & \textbf{3.0} & \textbf{3.7} & \underline{2.7} & \underline{3.5} \\ 
        mBART50-8k & \textbf{9.2} & \textbf{11.1} & \textbf{8.9} & \textbf{11.3} & 0.0 & 0.0 & \underline{2.8} & \underline{3.6} & \textbf{2.8} & \textbf{3.6} \\ 
        \hline
    \end{tabular}}

    \vspace{8px}
    \scalebox{0.9}{
    \begin{tabular}{lcccccccccc}
        \hline
        \multirow{2}{*}{\textbf{Model}} & \multicolumn{2}{c}{\textbf{Lithuanian}} & \multicolumn{2}{c}{\textbf{Maltese\,*}} & \multicolumn{2}{c}{\textbf{Polish}} & \multicolumn{2}{c}{\textbf{Portuguese}} & \multicolumn{2}{c}{\textbf{Romanian}} \\
         & \textit{F1@5} & \textit{F1@M} & \textit{F1@5} & \textit{F1@M} & \textit{F1@5} & \textit{F1@M} & \textit{F1@5} & \textit{F1@M} & \textit{F1@5} & \textit{F1@M} \\
        \hline
        YAKE & 0.0 & 0.0 & 0.0 & 0.0 & 0.0 & 0.0 & 0.0 & 0.0 & 0.0 & 0.0 \\ 
        mT5-small & 1.1 & 1.9 & 3.2 & 4.9 & 3.6 & 5.4 & 2.5 & 4.1 & 1.4 & 2.4 \\ 
        mT5-base & 2.0 & 3.4 & 3.9 & 6.1 & 5.0 & 7.5 & 3.5 & 5.7 & 2.4 & 4.0 \\ 
        mT5-large & 2.1 & 3.5 & 4.0 & 6.3 & 4.9 & 7.2 & 3.3 & 5.4 & 2.5 & 4.1 \\
        mBART50 & \underline{3.1} & \underline{4.2} & \textbf{6.0} & \textbf{7.7} & \underline{9.9} & \underline{11.7} & \underline{4.4} & \underline{5.5} & \underline{4.2} & \underline{5.4} \\ 
        mBART50-8k & \textbf{3.5} & \textbf{4.7} & \underline{5.4} & \underline{7.1} & \textbf{11.2} & \textbf{13.2} & \textbf{5.1} & \textbf{6.5} & \textbf{4.4} & \textbf{5.7} \\
        \hline
    \end{tabular}}

    \vspace{8px}
    \scalebox{0.9}{
    \begin{tabular}{lcccccccc}
        \hline
        \multirow{2}{*}{\textbf{Model}} & \multicolumn{2}{c}{\textbf{Slovak\,*}}& \multicolumn{2}{c}{\textbf{Slovene}} & \multicolumn{2}{c}{\textbf{Spanish}} & \multicolumn{2}{c}{\textbf{Swedish}} \\
         & \textit{F1@5} & \textit{F1@M} & \textit{F1@5} & \textit{F1@M} & \textit{F1@5} & \textit{F1@M} & \textit{F1@5} & \textit{F1@M} \\
        \hline
        YAKE & 0.0 & 0.0 & 0.0 & 0.0 & 0.0 & 0.0 & 0.0 & 0.0 \\ 
        mT5-small & 6.2 & 9.8 & 4.9 & 8.3 & 1.4 & 2.4 & 1.1 & 1.9 \\ 
        mT5-base & 6.8 & 10.7 & 5.6 & 9.4 & 2.1 & 3.5 & 2.3 & 3.9 \\ 
        mT5-large & 7.0 & 11.0 & 6.1 & 10.3 & 2.2 & 3.6 & 2.0 & 3.3 \\ 
        mBART50 & \underline{12.8} & \underline{15.7} & \underline{8.4} & \underline{11.1} & \underline{3.7} & \textbf{4.8} & \underline{3.3} & \underline{4.2} \\ 
        mBART50-8k & \textbf{13.0} & \textbf{16.0} & \textbf{8.6} & \textbf{11.8} & \textbf{3.6} & \textbf{4.8} & \textbf{3.7} & \textbf{5.0} \\ 
        \hline
    \end{tabular}}
    \caption{\textbf{Absent} keyphrases prediction results for the multilingual setting. Languages with a star (\textasteriskcentered{}) are unsupported by mBART50. Croatian with a dagger (\textdagger) is unsupported by mT5.}
    \label{tab:detailed_scores_absent_kps}
\end{table*}

\clearpage

\section{Random Split}
\label{Appendix:RandomSplit}
As discussed previously, we perform a temporal split on the data to better reflect the reality of temporal concept shifting as well as new languages being introduced in the EU. 
However, a random split might be better in some cases, such as if we train a model for short-time use, and are willing to recollect data regularly to insure performance remains as predicted. 
In short, a random split can only procure guarantee about its performance as long as the real world data remains similar to the training data, while a temporal split give a view of the performance while accounting for changes in the real world data.

Still, to better allow practitioners to use data as they desire, we released a random split (\url{https://huggingface.co/datasets/NCube/europa-random-split}).
The training, validation and test sets have respectively 10\,003, 3\,391 and 4\,439 judgments.
These figures are the same as in the chronologically split dataset.
However, the number of instances per set in the random split setting differs since each set has the same distribution in terms of languages.
Consequently, each set contains 159\,306, 53\,943 and 71\,708 instances respectively.
The results achieved by mBART50 are shown in Table~\ref{tab:Append:ResultsRandomSplit}. 
As expected, results are overall much higher since all training, validation and testing sets have the same distribution in terms of languages and vocabulary. 
We want to emphasize once again however that due to the fast changing nature of legal NLP, these performances are only valid for a short time after the data collection.

\vfill
\begin{table}[h]
\centering
\resizebox{0.8\textwidth}{!}{
\begin{tabular}{lccccccc}
\hline
\multirow{2}{*}{\textbf{Language}} & \multicolumn{3}{c}{\textbf{F1 Present}} & \multicolumn{3}{c}{\textbf{F1 Absent}} & \textbf{MAP} \\
 & \textit{@5} & \textit{@10} & \textit{@M} & \textit{@5} & \textit{@10} & \textit{@M} & \textit{@50} \\
\hline
Weighted Avg. & 30.4 & 20.0 & 41.5 & 15.3 & 10.3 & 18.1 & 20.6\\
Unweighted Avg. & 30.6 & 20.2 & 41.3 & 15.2 & 10.3 & 17.9 & 20.7\\
\hline
French & 30.2 & 19.7 & 41.2 & 14.9 & 9.8 & 17.8 & 20.2\\
German & 30.0 & 19.8 & 40.3 & 14.2 & 9.3 & 18.4 & 19.8\\
English & 32.8 & 22.0 & 43.6 & 9.9 & 6.5 & 13.3 & 19.0\\
Italian & 27.7 & 18.2 & 38.7 & 16.3 & 10.7 & 20.0 & 19.4\\
Dutch & 31.1 & 20.2 & 43.0 & 12.9 & 8.5 & 15.9 & 18.9\\
Greek & 12.6 & 7.8 & 21.7 & 15.9 & 11.1 & 17.9 & 12.0\\
Danish & 27.8 & 18.3 & 37.5 & 13.9 & 9.3 & 18.0 & 18.2\\
Portuguese & 27.4 & 17.7 & 39.5 & 14.5 & 9.6 & 16.9 & 17.8\\
Spanish & 30.1 & 19.7 & 41.2 & 14.1 & 9.2 & 17.2 & 19.2\\
Swedish & 34.8 & 23.4 & 44.7 & 10.5 & 7.1 & 13.7 & 21.1\\
Finnish & 28.2 & 17.7 & 42.4 & 17.9 & 12.3 & 21.0 & 20.1\\
Lithuanian & 38.5 & 26.0 & 49.0 & 12.1 & 8.2 & 14.8 & 24.0\\
Estonian & 40.2 & 28.0 & 48.1 & 9.8 & 6.6 & 12.4 & 24.8\\
Czech & 28.9 & 18.6 & 41.8 & 24.4 & 16.9 & 26.6 & 24.5\\
Hungarian & 27.1 & 17.2 & 40.3 & 18.4 & 12.5 & 21.6 & 20.1\\
Latvian & 38.7 & 26.2 & 48.8 & 11.4 & 7.7 & 14.1 & 23.4\\
Slovenian & 33.4 & 22.6 & 43.3 & 17.6 & 11.9 & 21.1 & 24.0\\
Polish & 27.4 & 17.2 & 41.0 & 24.3 & 17.4 & 25.6 & 24.3\\
Maltese & 28.1 & 17.9 & 40.5 & 18.3 & 12.6 & 20.5 & 19.8\\
Slovak & 31.3 & 20.0 & 45.0 & 22.0 & 15.2 & 24.0 & 24.2\\
Romanian & 35.2 & 23.5 & 46.5 & 19.4 & 13.2 & 21.5 & 25.6\\
Bulgarian & 35.4 & 23.8 & 46.3 & 11.2 & 7.6 & 13.1 & 21.9\\
Croatian & 40.8 & 28.6 & 46.9 & 13.7 & 9.3 & 13.9 & 26.4\\
Irish & 15.9 & 11.5 & 20.5 & 7.7 & 5.1 & 10.2 & 8.9 \\
\hline
\end{tabular}}
\caption{Results on a random split of the data, per language, achieved by mBART50.}
\label{tab:Append:ResultsRandomSplit}
\end{table}
\vfill

\end{document}